\documentclass[a4paper,conference]{IEEEtran}

\usepackage{multirow}

\usepackage{cite}
\usepackage{hyperref}
\usepackage{booktabs}
\usepackage{amssymb}
\usepackage{xurl}
\usepackage{mwe}
\usepackage[dvipsnames,table]{xcolor}
\usepackage{tikz}
\usetikzlibrary{backgrounds}
\usetikzlibrary{arrows,shapes}
\usetikzlibrary{tikzmark}
\usetikzlibrary{calc}
\usepackage{graphicx}
\usepackage{amsmath}
\usepackage{algorithmicx}
\usepackage{algpseudocode}
\usepackage{array}
 \usepackage{mwe}
 \usepackage{tabularx}
 \usepackage{nicematrix}

\usepackage{tcolorbox}

\newcolumntype{Y}{>{\centering\arraybackslash}X}
\usepackage{tikz}
\usetikzlibrary{arrows,shapes,positioning,shadows,trees,mindmap}
\usepackage[edges]{forest}
\usetikzlibrary{arrows.meta}
\colorlet{linecol}{black!75}
\usepackage{xkcdcolors} % xkcd colors

\usepackage{tikz}
\usetikzlibrary{backgrounds}
\usetikzlibrary{arrows,shapes}
\usetikzlibrary{tikzmark}
\usetikzlibrary{calc}
\newcommand{\highlight}[2]{\colorbox{#1!17}{$\displaystyle #2$}}
\colorlet{mhpurple}{Plum!80}

\renewcommand{\highlight}[2]{\colorbox{#1!17}{#2}}

\ifCLASSOPTIONcompsoc
 \usepackage[caption=false,font=normalsize,labelfont=sf,textfont=sf]{subfig}
\else
 \usepackage[caption=false,font=footnotesize]{subfig}
\fi
\usepackage{stfloats}
\usepackage{url}

\newcommand{\myParagraph}[1]{
\noindent \textbf{#1} ---
}

\begin{document}
% at, but, by, for, in, nor, of, on, or, the, to and up
\title{MoCap-less Quantitative Evaluation of Ego-Pose Estimation Without Ground Truth Measurements}

% authors
\author{\IEEEauthorblockN{Quentin Possamaï\IEEEauthorrefmark{1}\IEEEauthorrefmark{2},
Steeven Janny\IEEEauthorrefmark{1}\IEEEauthorrefmark{2},
Guillaume Bono\IEEEauthorrefmark{1},
Madiha Nadri\IEEEauthorrefmark{2},
Laurent Bako\IEEEauthorrefmark{3} and
Christian Wolf\IEEEauthorrefmark{1}}
\IEEEauthorblockA{\IEEEauthorrefmark{1}LIRIS, INSA-Lyon, CNRS, Villeurbanne, France}
\IEEEauthorblockA{\IEEEauthorrefmark{2}LAGEPP, Université Claude Bernard, Villeurbanne, France}
\IEEEauthorblockA{\IEEEauthorrefmark{3}AMPERE, Centrale Lyon, Ecully, France}}

\maketitle

\begin{abstract}
The emergence of data-driven approaches for control and planning in robotics have highlighted the need for developing experimental robotic platforms for data collection. However, their implementation is often complex and expensive, in particular for flying and terrestrial robots where the precise estimation of the position requires motion capture devices (MoCap) or Lidar. In order to simplify the use of a robotic platform dedicated to research on a wide range of indoor and outdoor environments, we present a data validation tool for ego-pose estimation that does not require any equipment other than the on-board camera. The method and tool allow a rapid, visual and quantitative evaluation of the quality of ego-pose sensors and are sensitive to different sources of flaws in the acquisition chain, ranging from desynchronization of the sensor flows to misevaluation of the geometric parameters of the robotic platform. 
Using computer vision, the information from the sensors is used to calculate the motion of a semantic scene point through its projection to the 2D image space of the on-board camera. The deviations of these keypoints from references created with a semi-automatic tool allow rapid and simple quality assessment of the data collected on the platform. To demonstrate the performance of our method, we evaluate it on two challenging standard UAV datasets as well as one dataset taken from a terrestrial robot.
\end{abstract}

\IEEEpeerreviewmaketitle

\section{Introduction}
% Big problem : robotics, control and stuff
\noindent
The deployment of robotic solutions 
%has became prominent in various application from industrial automation of supply chains to private home assistant by way of unmanned aerial vehicles for deliveries. 
%Going through such expansion requires reliable and robust control and planning algorithms to provide off-the-shelf robots for real world application. To design such algorithm, the domain have recently transition from model-based approaches, which consists in analytical models of the robot actuation and sensors dynamics, to data-driven approaches, which allows to deeply integrate computer vision into the control pipeline.
is an important goal, 
yet research and development in the field is hampered by difficulties setting up hardware and software to build and operate a fully instrumented robotic platform. Such platforms often require expensive sensors and complex libraries for accurate measurement, in particular odometry (vehicle ego-pose estimation), usually estimated through inertial measurement units (IMUs), Lidar, or visual observations captured from on-board cameras. A large amount of engineering effort is required to calibrate sensors, synchronize them and to optimize the estimation algorithms, which are often based on multi-modal fusion. Although downstream tasks strongly rely on high data quality \cite{AssemExpStudyArxiv2021}, evaluation and validation is not straight forward, as ground truth (GT) data on vehicle motion is difficult to come by. Standard algorithms used to obtain GT data usually rely on marker-based motion capture (MoCap), or, less precise, with 2D QR-code-like markers observed with RGB cameras (eg. ARToolkit\footnote{\url{https://github.com/artoolkit}}). Both of these methods are limited to indoor scenes, at least for realistic settings. Evaluating and validating odometry in difficult outdoor conditions remain challenging issues, in particular for UAV operations, where the vehicle trajectory can evolve far from any base platform equipment. On the other hand, abandoning validation all-together and blindly relying on the quality of the equipment without proof can turn out to be costly, as late discovery of dramatic flaws in large amounts of capture data often requires expensive reboot of the project.

%and many projects discover dramatic flaws in large amounts of capture data, requiring an expensive reboot of the project.

\begin{figure}[t] \centering
\includegraphics[width=\columnwidth]{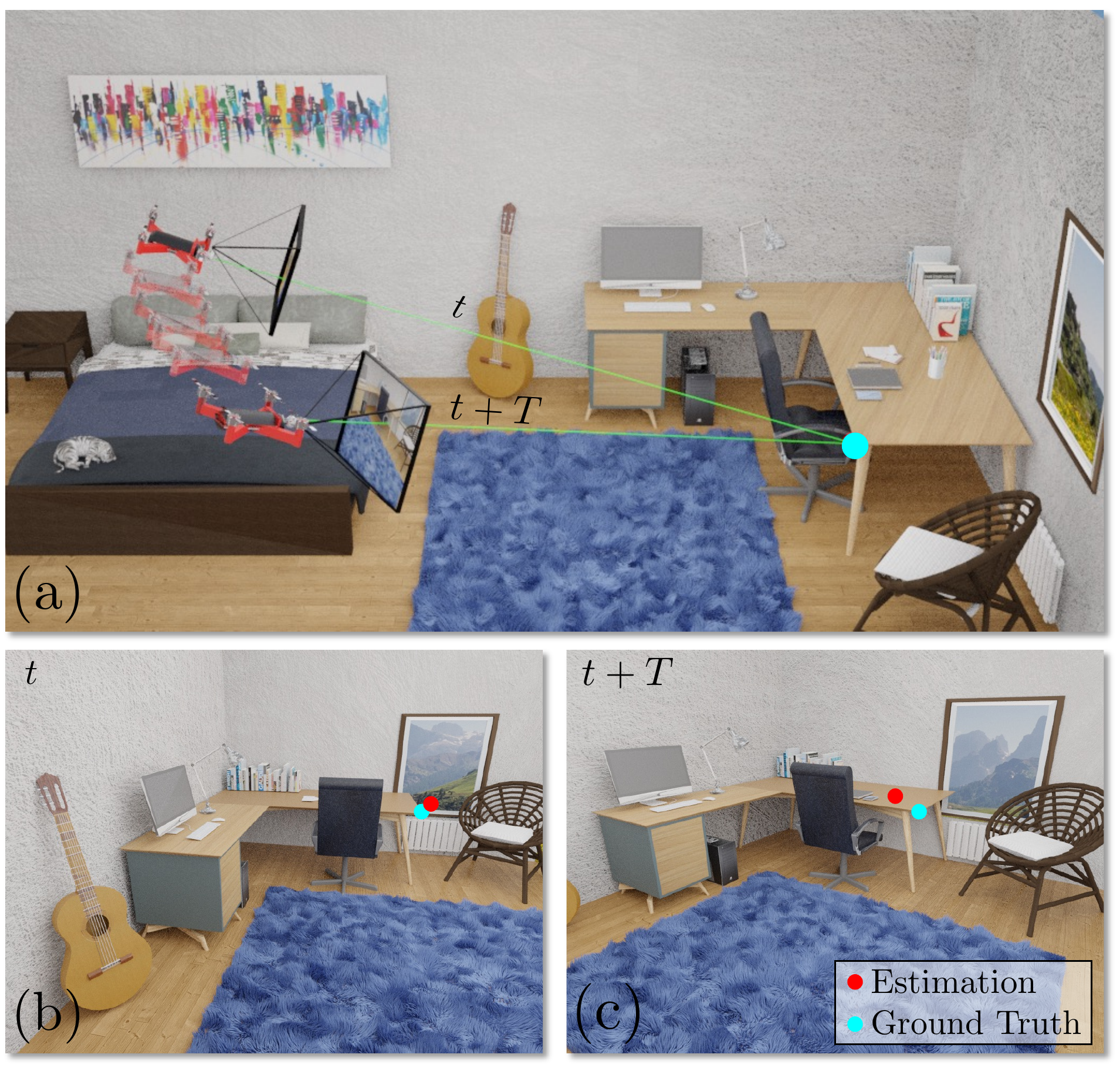}
\caption{\label{fig:teaser}(a) We evaluate any odometry/ego-pose estimation system, for instance based on inertial measurement units (IMUs), by tracking the points painted by a virtual laser mounted on the agent. (b) The virtual laser position is initialized randomly at the start of an episode (frame $t$) and painted in the observed RGB image. (c) At frame $t{+}T$, the laser direction is updated from estimated odometry and the point position updated. Sensor quality is assessed by measuring the pixel-wise distance with respect to the ground truth keypoint position obtained semi-interactively.}
\vspace{-5mm}
\end{figure}

% Narrower problem : on board sensors requires post-processing and synchronization, data quality are fundamental for many application, including model identification, sensors fusion for planing, etc... even reinforced by the development of data-base approach

% Yet Narrower paper gap : Evaluate quality of data flow measured by different sensors on a robotics platform
In this paper, we address the crucial task of odometry data validation for vision-based robotic platforms, an important problem for terrestrial or UAV based navigation platforms. We propose a tool to rapidly estimate errors in the data gathering pipeline and evaluate the precision of the measurements. Our proposed method is semi-automatic and based on the following observation: (i) precise manual annotation of vehicle ego-motion is extremely difficult for humans, (ii) annotating semantically meaningful points in a scene and tracking them is easy and is performed as a common task in computer vision, in particular for GT creation for supervised machine learning. 

Starting from this idea, we translate vehicle ego-motion into motion of keypoints in the 2D image space of the observed RGB frames captured by an on-board camera, as shown in Figure \ref{fig:teaser}.
The software chooses a point $\mathbf{p}^\mathcal{I}$ in the 2D image frame observed by the on-board camera, and draws it as a red dot superimposed on the image. The actual choice of point is not important and can be arbitrary set; however, we suppose the existence of a 3D point in the scene $\mathbf{p}^\mathcal{C}$, eg. a point on a bush or shrub, a corner of table etc. This scene point is the point selected by passing a ray through the optical center of the camera and through $\mathbf{p}^\mathcal{I}$. We are interested in the estimation of the ego-motion occurring after some time $T$, at frame $t+T$. The  point $\mathbf{p}^\mathcal{C}$ has not moved, but the point $\mathbf{p}^\mathcal{I}$ in general has moved in the image if ego-motion did occur. We use the estimated motion information to move the colored keypoint to a new 2D position and superimpose it on the RGB image observed at instant $t+T$. The $(x,y$) coordinates in 2D image space of the keypoints at time instants 
$t$ and $t+T$, respectively, will generally be different; however, if the ego-motion information delivered by the sensors (and to be validated by our method) is precise, then the two keypoints will be superimposed on \textit{the same semantic point in the scene}. This is illustrated in Figure \ref{fig:teaser}: in the two frames (b) and (c) the red dot is on different $(x,y)$ coordinates in image space, but it falls on the same spot in the scene, namely on one of the corners of the desk. We leverage a semi-automatic tracking tool and human intervention to estimate this error and translate it into validation of ego-motion itself.

The main difficulty in this geometric approach lies in the absence of depth information, which is often hard to obtain in challenging outdoor scenarios, and which makes inverse geometric projections difficult. Our \textit{Ansatz} solves this problem in two different and alternative ways: either through constraints on the starting configuration, or, if available, leveraging prior information on the scene geometry itself.

%uses 2D projection of 3D poses estimation to quantitatively evaluate the overall quality of the data gathering including (a) time synchronisation issues between sensors (b) kinematics misevaluation, and (c) sensor noises. By getting rid of the need for expensive ground truth labelling frameworks, our tool can help researchers to validate the we claim the followingiroboticsly optimize the higher level research on planning and control algorithms.

To summarize, we claim the following contributions: 
\begin{itemize}
    \item We introduce VICE (\textit{Visual Inertial Correlation Estimator}), a method and tool to intuitively and quantitatively evaluate quality of data produced by robot ego-motion sensors and algorithms.
    \item We evaluate the method on three heterogenous platforms including slow and agressive UAV maneuvers and terrestrial navigation. We compare the error estimated by VICE with the error provided by MoCap.
\end{itemize}

%Camera and Inertial Measurement Unit or IMU are very popular sensor among nearly-all moving systems such as UAV. Moreover visual-inertial odometry methods are popular\cite{nisar_vimo_2019}. However on the field in is important that the sensors are calibrated and synchronised and that they operate correctly. 

%Traditionally, motion estimation techniques are often evaluated and calibrated with external high-precision sensors like marker based motion capture equipment. However, this equipment is mostly only applicable in smaller indoor environments. Estimating the precision of ego-motion sensors in the wild in out door conditions is a non-trivial task.

%To evaluate qualitatively and quantitatively we propose a visual inertial correlation estimator that will produce an visual and numerical error between one camera and one 6DoF pose estimate....

\section{Related work}

\myParagraph{Odometry and ego-motion} is traditionally estimated with dedicated sensors like encoders on wheels for terrestrial robots or IMUs for drones. Additionally, visual information can provide complementary cues (``visual odometry'') either through methods detecting local features (keypoints)  and matching \cite{revaud_r2d2_2019}, \cite{geneva_openvins_2020} or dense methods, often based on learning \cite{liu_tlio_2020},  \cite{nisar_vimo_2019},  \cite{chen_ionet_2018}.

\myParagraph{SLAM (Simultaneous Localization and Mapping)} leverages motion estimation into a method which creates a map on the fly and positions an agent on it \cite{cadena_simultaneous_2016}. SLAM methods can thus also be used for ego-motion estimation, as each new observation can be aligned with the map. Different variants exist depending on the type of input (Lidar, RGB, RGB-D), the type of primitives used for matching (e.g. dense \cite{engel_lsd-slam_2014} or keypoint based \cite{mur-artal_orb-slam_2015}) and the type of map generated (2D \cite{SLAMROS2021} or 3D metric \cite{newcombe_kinectfusion_2011}\cite{engel_lsd-slam_2014}, topological \cite{TopoSLAM2001} etc.). Links also exist to structure from motion, a similar problem in computer vision \cite{forsyth_computer_2011}. Another alternative to odometry or SLAM useful for ego-motion estimation is to go through absolute positioning, i.e. camera localization, using similar keypoint or dense techniques \cite{revaud_r2d2_2019}\cite{SuperGLUECVPR202}.

\myParagraph{Sensor synchronization} is ideally done within the hardware. However it is not always possible, especially between external and internal sensors. Signal analysis can be required to estimate the synchronization offset using cross-correlation\cite{fridman_automated_2015}, \cite{ploetz_automatic_2012}, \cite{zhang_syncwise_2020}.

\myParagraph{Calibrating and evaluating motion estimation} allows the exploitation of the measures of a vast variety of sensors. The most notable one is the IMU that is very noisy and nearly always requires a calibration \cite{rehder_extending_2016} or a de-noising \cite{brossard_denoising_2020}. After their integration into an ego-pose, the position and the orientation need to be evaluated. While the position is generally easily managed, the orientation has its challenges as some forms have non continuous properties. \cite{geneva_openvins_2020} proposes an absolute metric while \cite{zhang_tutorial_2018} defines a relative to the former.

\section{Semi-automatic motion estimation}
\noindent 
%With VICE, we evaluate the general accuracy of any odometry algorithm by translating the complex problem of validating the 3D position of a vehicle into a simpler problem of tracking 2D points in image space. A quantitative evaluation of the ego-pose is obtained by computing the difference between the result of the tracking process (based on ego-pose measurements), and the exact position of a semantic object located by a human operator. In a nutshell, our approach is as follows:
We introduce VICE, a semi-automatic procedure for odometry validation on computer-vision oriented robotic platform. Our method performs by tracking displacement of several reference points forward in time using the ego-pose estimate. These references are compared to manually annotated keypoints (using our annotation tool\footnote{\url{https://github.com/quentinpossamai/vice}}) marking a fixed semantic point.

%VICE is a method for both quantitative and qualitative assessment of the overall quality of an odometry solution. Our tool does not require accurate labels of 3D pose from high-end external sensors (MoCap, Lidar, etc...) and is solely based on the image flow from an on-board camera on the mobile robot. To do so, we compare the position of static keypoints in the scene projected onto the 2D image space with the estimation of their positions obtained thanks to the ego-pose measurements. Thus, a small distance between these two measurements will indicate the good performance of the odometry. In a nutshell, our approach is as follows:

% \begin{enumerate}
%     \item During a \textbf{human labelling} step, an operator marks the 2D position of one or more semantic reference points on the video extracted from the on-board camera. This step may be performed manually using an annotation tool\footnote{\url{https://github.com/quentinpossamai/annotation_tool}}.
%     \item A \textbf{keypoints tracking} algorithm is deployed to extrapolate 3D positions of the reference keypoints. This step make use of the odometry measurements.
%     \item Finally, \textbf{qualitative \& quantitative assessment} are performed by numerically measuring the difference between the 2D position of the reference keypoints marked in step 1 and their estimate based on the robot movements calculated in step 2. A qualitative analysis is also conducted by plotting on the video both the reference point and its estimate.
% \end{enumerate}

\subsection{Geometric formalization of the problem}

\noindent
We first formally introduce the different geometrical coordinate systems and transformations. 
Let $\mathcal{B}$ the local coordinate system of the agent, which can be either identical to the reference frame of an IMU, or correspond to the center of mass of the robot;
%if we use the method to evaluate the precision of an IMU
 $\mathcal{F}$ the fixed global coordinate system in which the ego-pose is estimated, and which we set without loss of generality to be equal to $\mathcal{B}$ at instant $t{=}0$;  let $\mathcal{C}$ be the 3D coordinate system of the camera, which is in general not equal to $\mathcal{B}$; finally, let $\mathcal{I}$ be the 2D image plane of the camera. We assume that the transform $T^{\mathcal{B},\mathcal{C}}$ is known, but not necessarily constant in time. 
%Note the the 3D reference point is expressed in $\mathcal{C}$ after applying the 2D to 3D conversion. 
We suppose the camera to be calibrated, i.e. camera intrinsics are known and composed of
\begin{itemize}
    \item the camera matrix:
\[\mathbf{K} = \begin{bmatrix}
f_x & 0 & c_x \\
0 & f_y & c_y \\
0 & 0 & 1 
\end{bmatrix}\]
with $f_x, f_y$ the focal length and $c_x, c_y$ the optical centers, 

\item the radial distortion coefficients $k_i, {i=1,6} $ and tangential distortion coefficients $p_1, p_2$ modelled in an image distortion function $\zeta(.)$ \cite{Brown1966DecenteringDO}.
\end{itemize}
Projection and inverse projection to and from the image plane $\mathcal{I}$ require depth values for each point, which can be obtained with a sensor (z-Cam, stereo). In our experiments we also addressed the case where the robot is \emph{not} equipped with depth sensors and propose a different way to recover depth, described in section \ref{sec:estimatingdepth}.

The task itself can be described as follows: given a trajectory of the agent, 
we are given a sequence of relative pose estimates $T^{\mathcal{B}_{t+1}, \mathcal{B}_t}$ $t \in [0, n-1]$ with $n$ the length of the trajectory. These pose estimates correspond to the motion of the agent between two time instants in the agent coordinate frame $\mathcal{B}$ and the \textit{evaluation of their precision is the goal of this work}. These estimates may be obtained directly as relative motion between two instants, i.e. with an IMU or through visual odometry, or in an absolute coordinate frame with respect to a global reference, for instance in the case of visual camera localization methods \cite{revaud_r2d2_2019}, \cite{SuperGLUECVPR202}. Without loss of generality, we suppose that a relative transformation between two time instants is measured. The pose $T^{\mathcal{B}_{t+1}, \mathcal{B}_t}$ is thus the transformation matrix w.r.t $\mathcal{B}_{t+1}$ describing $\mathcal{B}_{t}$,

\[T^{\mathcal{B}_{t+1}, \mathcal{B}_t} = \begin{bmatrix}
R^{\mathcal{B}_{t+1}, \mathcal{B}_t} & d^{\mathcal{B}_{t+1}, \mathcal{B}_t} \\
0 & 1
\end{bmatrix}\]
with $R^{\mathcal{B}_{t+1}, \mathcal{B}_t}$ the rotation matrix and $d^{\mathcal{B}_{t+1}, \mathcal{B}_t}$ the translation vector.

Additionally, %and synchronized with these measurements, 
we are given images $I_t$ from the on-board camera. Image pixel coordinates $\mathbf{p}^\mathcal{I}=(u,v)$ are expressed in image plane $\mathcal{I}$. 
%Our method do not require precise synchronization between images and ego-pose estimation, as VICE is designed to detect such temporal misalignment.

\begin{figure}[t] \centering
    \includegraphics[width=\linewidth]{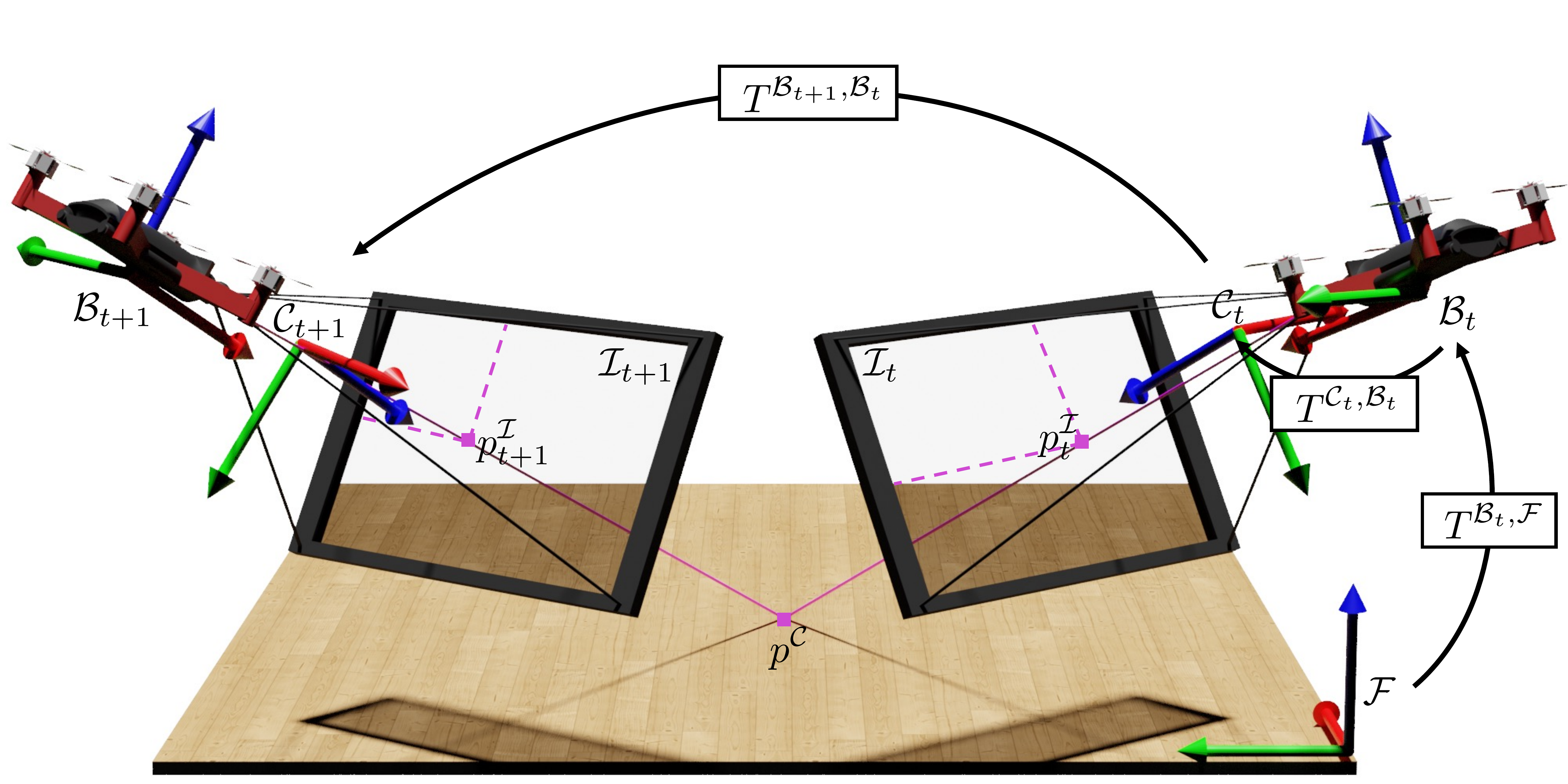}
    \caption{Our method takes a fixed scene point and tracks its projection to $\mathbf{p}^\mathcal{I}$ the 2D image space  over time. 
    We leverage ego-pose estimation $T^{\mathcal{B}_{t+1},\mathcal{B}_t}$ to track the displacements.}
    \label{fig:point_tracking}
    \vspace{-5mm}
\end{figure}

\subsection{Tracking ego-motion as keypoint motion}

\noindent
We translate ego-motion (to be evaluated) into the motion of an observed keypoint in the onboard images as follows: at the beginning of a trajectory at instant $t=0$, a keypoint  $\mathbf{p}^\mathcal{I}_t$ is chosen in the observed frame $I_t$. % randomly, w.l.o.g. It will be taken in the center of the image. 
This keypoint corresponds to a semantic point in the scene $\mathbf{p}^\mathcal{C}_t$, e.g. a point on an object, a surface, or a tree, and the objective is to track the corresponding point in the scene over the next observed images of the sequence, which requires the correct estimation of camera motion, linked to ego-motion (see Fig. \ref{fig:point_tracking}).

More generally, this tracking process of the same semantic scene point across successive images involves the following steps: (a) inverse perspective projection of $\mathbf{p}^\mathcal{I}_{t=0}$ from  2D image space to 3D camera space $\mathcal{C}$ and change to the agent coordinate system $\mathcal{B}$; (b) application of the measured (and to be evaluated) ego-motion $T^{\mathcal{B}_{t+1}, \mathcal{B}_t}$; (c) change back to the camera coordinate system $\mathcal{C}$ and perspective projection to the image frame $\mathcal{I}$, obtaining $\mathbf{p}^\mathcal{I}_{T}$. The full transformation can be given as

\vspace{-7mm}
\begin{equation}
\label{eq:tracking}
\mathbf{p}^\mathcal{I}_T = 
\tikzmarknode{tj1}{\highlight{NavyBlue}{\color{black}  $\phi T^{\mathcal{C}, \mathcal{B}_T}$}
    }
\color{purple}
\overbrace{ 
    \color{black}
    \left(\prod_{t=1}^T T^{\mathcal{B}_{t}, \mathcal{B}_{t-1}}\right)
}^{\substack{\text{\sf \footnotesize \textcolor{purple!85}{Forward transform}} 
          \\ \text{\sf \footnotesize \textcolor{purple!85}{in time}} } }
\color{black}
\tikzmarknode{tj}{\highlight{Bittersweet}{ \color{black} $T^{\mathcal{B}_0, \mathcal{C}}\phi^{-1}(\mathbf{p}^\mathcal{I}_0, z)$}}
\end{equation}
\vspace*{0.5\baselineskip}
\begin{tikzpicture}[overlay,remember picture,>=stealth,nodes={align=left,inner ysep=1pt},<-]
\path (tj1.north) ++ (-1.8,-2em) node[anchor=north west,color=NavyBlue!85] (tj1text){$\substack{\text{\sf \footnotesize projection back to} \\ \text{\sf \footnotesize image frame } \mathcal{I}_T}$};
\draw [color=NavyBlue]([xshift=6mm]tj1.south) |- ([xshift=-0.3ex,color=NavyBlue]tj1text.south west);
% For "t_{j}"
\path (tj.north) ++ (-0.3,-2em) node[anchor=north west,color=Bittersweet!85] (tjtext){$\substack{{\textsf{\footnotesize Reference point } } \\  {\textsf{\footnotesize in $\mathcal{B}$ at } t=0 }}$};
\draw [color=Bittersweet]([xshift=-1em]tj.south) |- ([xshift=-0.3ex,color=Bittersweet]tjtext.south east);
\end{tikzpicture}
\vspace{1mm}

\noindent
The transformation between the camera frame $\mathcal{C}$ and the agent $\mathcal{B}_T$ is supposed to be known, and can often be measured physically if access to the equipment is available. The projection operator $\phi$ %transformation $T^{\mathcal{I}\to \mathcal{C}}$ 
is a perspective projection to image space given the camera intrinsics $\mathbf{K}$ and $\zeta(.)$
\begin{equation}
\mathbf{p^\mathcal{I}} = \phi \left(\mathbf{p}^\mathcal{C}\right) = \zeta\left(\mathbf{\mathbf{K}p^\mathcal{C}}\right).
\end{equation}
Note that the inverse mapping $\phi^{-1}$ requires depth information ($z$ in Equation \eqref{eq:tracking}) which will be addressed in section \ref{sec:estimatingdepth}.  If all transformations are accurate, in particular if ego-motion $T^{\mathcal{B}_{t+1},\mathcal{B}_t}$ has been correctly estimated, then
the two points $\mathbf{p}^\mathcal{I}_{T}$ and $\mathbf{p}^{\mathcal{I}}_0$ in the two respective images $I_{T}$ and $I_{0}$ will fall on the same semantic scene point, although in general their 2D image coordinates will differ.

\subsection{Comparing ego-motion with reference annotations}

\noindent
Since the quality of the ego-motion estimate cannot be measured by the distance between 2D keypoints $|| \mathbf{p}^\mathcal{I}_{t-1} - \mathbf{p}^\mathcal{I}_{t}||$, but rather by the difference in their respective (unknown) scene points, we use human annotation in 2D image space to provide ground truth (GT) measurements.
%through the following procedure.

\begin{figure}[t] \centering
\includegraphics[width=\columnwidth]{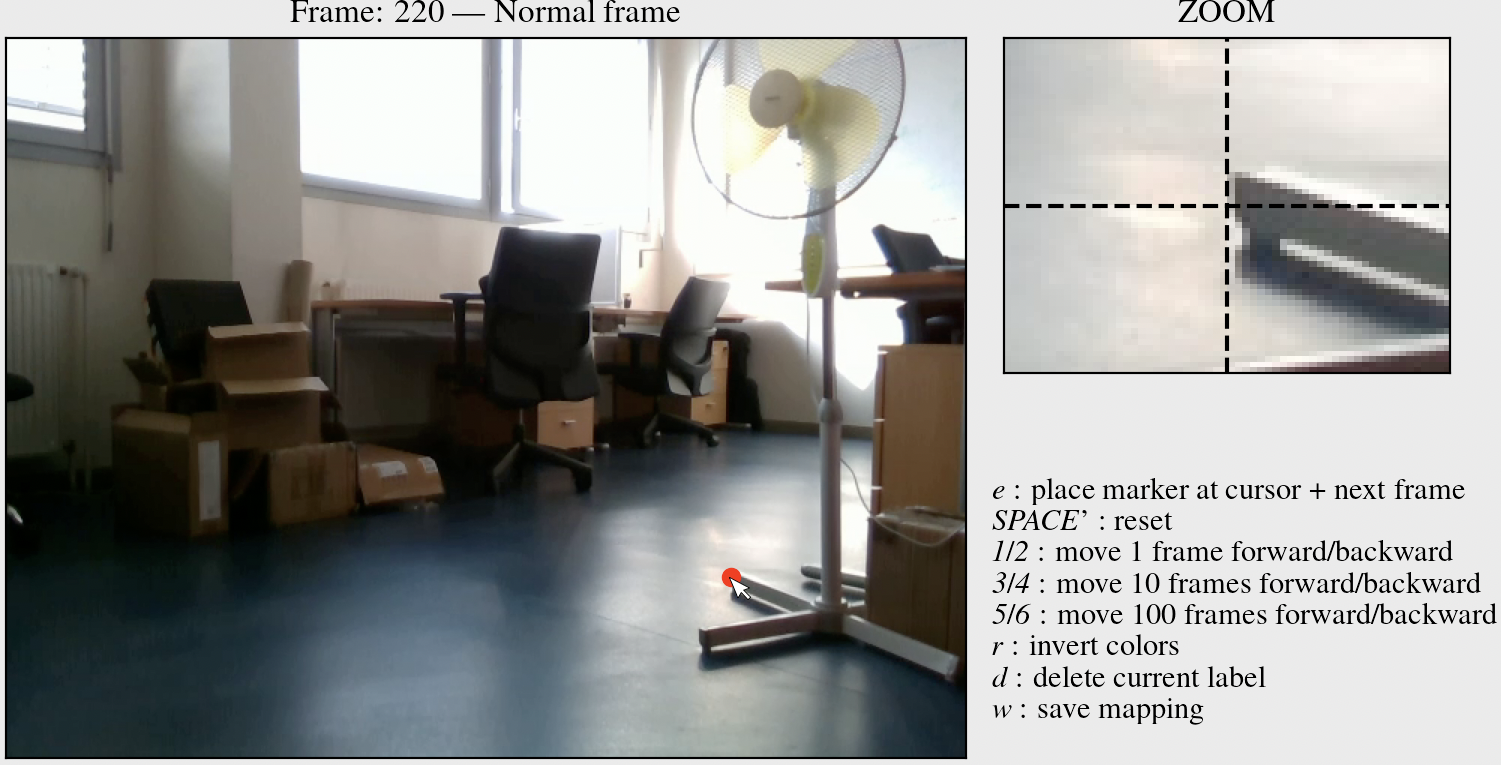}
\caption{\label{fig:interface}The semi-automatic annotation process allows a human to provide GT estimates for ego-motion through the annotation of semantically consistent 2D points in the scene.}

    \vspace{-5mm}
\end{figure}

For each trajectory, a video is created showing the observed camera frames $I_t$. At the beginning of a sequence, for frame $I_{t=0}$ a 2D keypoint position is initialized at $\mathbf{p}^\mathcal{I}_{t=0}$ and superimposed as a colored dot on the image. For the subsequent frames $I_t$, a human annotator is asked to annotate the GT positions  $\mathbf{p}^\mathcal{I*}_{t}$ which correspond to the same semantic scene point (see Figure \ref{fig:interface}). If the amount of camera motion leads to the point leaving the image space, a new reference point is spawned and annotation proceeds with this point as reference. The difference between $\mathbf{p}^\mathcal{I}_{t}$ (computed through the ego-motion estimated) and $\mathbf{p}^\mathcal{I*}_{t}$ (annotated) is used as a quality measure for the ego-motion estimate.

\subsection{Estimating depth}
\label{sec:estimatingdepth}

\begin{figure}[t] \centering
    \includegraphics[width=0.8\linewidth]{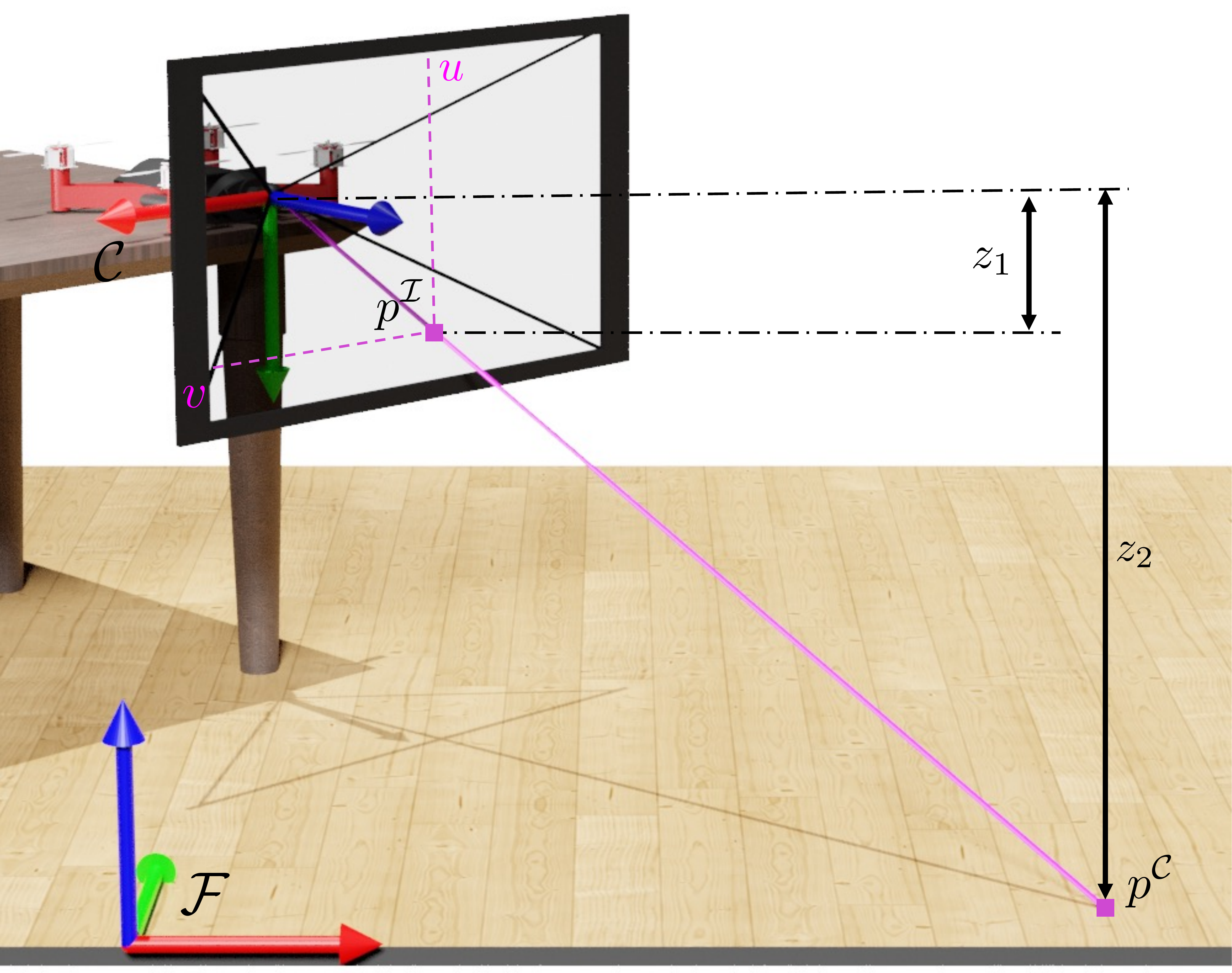}
    \caption{Obtaining depth values without direct measurements through a strong assumption on the scene geometry: we suppose the knowledge of some geometry information in the robot starting position, namely the translation and rotation between the agent camera frame $\mathcal{C}$ and a fixed global frame $\mathcal{F}$ that coincide with the floor. We leverage these prior measurements to virtually extend the ray from $p^\mathcal{I}$ to a point $p^\mathcal{C}$ coincident with the ground.
    %$(u, v)$ are the chosen pixels that will be projected in reference frame $C$.
    %$F$ is the fixed reference frame on the ground, , $z_1$ is defined as follow $(x_1, y_1, z_1)_F = R_{F, C} (x', y', 1)_C$.
    %$C$ is the camera reference frame, the point $(x', y', 1)_C$ is the projection of $(u, v)$ in $C$ in the camera plane (depth of 1), the point $\mathbf{p}^\mathcal{C}$ is the projection of $(x', y', 1)_C$ on the ground.
    }
    \label{fig:point_projection}
    \vspace{-5mm}
\end{figure}

\noindent 
Computing the inverse perspective projection from image frame $\mathcal{I}$ to the camera frame $\mathcal{C}$ requires the depth value for each considered image point, in particular the reference point at the first frame $I_0$, namely $\mathbf{p}^\mathcal{C}_{0} = \phi^{-1}(\mathbf{p}^\mathcal{I}_0, z)$. In cases where these values are not directly measured through an on-board device (z-cam or stereo rig), we propose a different approach to obtain them. By constraining the reference point to belong to a flat surface among the video, we circumvent the need for depth information. Formally, we bind $p^\mathcal{C}$ to the $(xy)$ plane define by the global frame $\mathcal{F}$ in which the ego-pose will ultimately be expressed. In most cases, this plane will coincide with the floor of the environment, and practically speaking, in this case this only requires the knowledge of the position of the camera w.r.t. to the floor at the beginning of the trajectory, as well as the camera angle. By leveraging this knowledge of the initial pose of the agent $T^{\mathcal{F},\mathcal{C}}_{t=0}$, we compute $\mathbf{p}^\mathcal{C}$ as follows:
\begin{equation}
    \mathbf{\tilde p}^\mathcal{C} = \phi^{-1}(\mathbf{p}^\mathcal{I}, 1),\quad \quad \mathbf{\tilde p}^\mathcal{F} = (x_1, y_1, z_1) = T^{\mathcal{F},\mathcal{C}}_{t=0} \mathbf{\tilde p}^\mathcal{C}
\end{equation}
\vspace{-5mm}
\begin{equation}
    \mathbf{p}^\mathcal{C} = \frac{z_2}{z_1} \frac{\mathbf{\tilde p}^\mathcal{C}}{\|\mathbf{\tilde p}^\mathcal{C}\|},
\end{equation}
where $z_2$ is the altitude of the camera with respect to the $(xy)$ plane of $\mathcal{F}$, and $z_1$ is the altitude gap between the camera optical center and the reference point on the camera plane (see fig. \ref{fig:point_projection}).  In the vast majority of cases, the fixed global frame $\mathcal{F}$ can be arbitrarily set, yielding easy measurement of the initial transformation $ T^{\mathcal{F},\mathcal{C}}_{t=0}$. In section \ref{sec:experiments}, we provide  numerical evidence that this solution can cope for the absence of depth sensor without being detrimental to the general evaluation of the overall odometry quality.

\subsection{Quantitative evaluation}
\noindent
Visual assessment is a crucial part of our method: it allows to rapidly evaluate the overall quality of the ego-pose estimates, and to identify specific flaws in the odometry algorithm such as camera-sensor de-synchronization, temporal drift or misevaluation of geometric parameters. Yet, we also propose quantitative metrics to rigorously compare different ego-pose estimates.

\par The most intuitive metric is certainly the pixel distance between the reference annotations $\mathbf{p}^\mathcal{I*}_t$ and the 2D points resulting from the tracking process, that is $\mathbf{p}^\mathcal{I}_t$ from \eqref{eq:tracking}. Formally:
\vspace{-3mm}
\begin{equation}
    \text{RMSE}_{2D} = \sqrt{\frac{1}{T}\sum_{t=1}^T \|\mathbf{p}^\mathcal{I*}_t - \mathbf{p}^\mathcal{I}_t\|^2_2}.
\label{eq:RMSE2D}    
\end{equation}
\noindent
Nevertheless, the limitation of VICE to a single point underlies an ambiguity concerning the evaluation of the ego-pose, in particular when the agent performs rotations and translations along the axis between the optical center of the camera and the position of the benchmark. To remove uncertainty, we recommend using more benchmarks and using VICE on all of them. Extensive experiments have been conducted on the impact of the number of reference keypoints, and our key results are presented in the following section.

\section{Experiments}
\label{sec:experiments}
\begin{figure*}
    \centering
    \includegraphics[width=\linewidth]{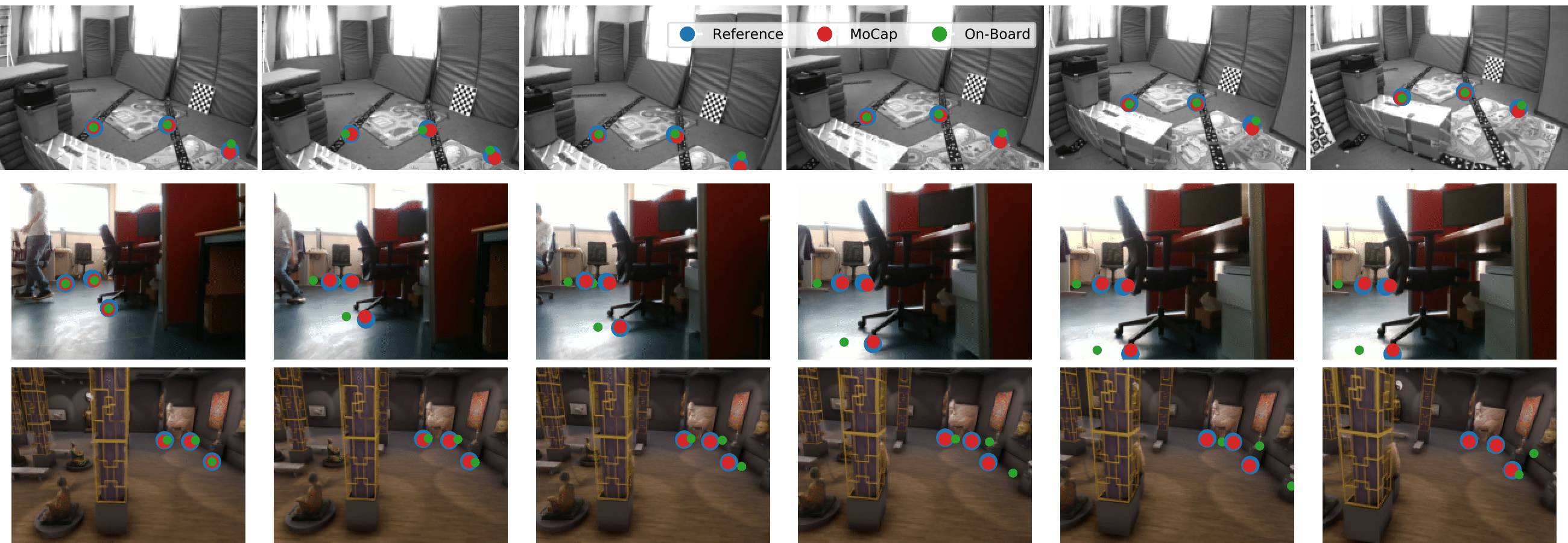}
    \caption{Qualitative evaluation on each dataset (from top to bottom, \textit{EuRoC MAV}, \textit{LoCoBot-INSAL} and \textit{Blackbird}). The observation clearly highlights the respective performances of the on-board sensors (in green) of each platform. We also notice that the MoCap (in red), although more precise, is also a noisy measure of the ego-pose, since the red points are not centered with respect to the references (in blue).}
    
    %\caption{Qualitative evaluation. Top row: \textit{EuRoC MAV} spaced with $0.5$ seconds; middle row \textit{LoCoBot-INSAL} spaced with $0.5$sec; bottom row: \textit{Blackbird} spaced with $0.33$sec. Cyan dot is the position of the label for all trajectories. Red dot: keypoint estimated by MoCap; Green dot: keypoint of onboard system to be evaluatd (IMU for \textit{EuRoC MAV} and \textit{Blackbird}; for \textit{LoCoBot-INSAL}, red dot is odometry and green cross is the SLAM).}
    \label{fig:3x7_vice}
    \vspace{-5mm}
\end{figure*}

\noindent We evaluated VICE's ability to assess ego-motion estimation algorithms on three different datasets
and several pose estimation algorithms. 
% More precisely, we answer the following question:
% Does VICE make it possible to qualitatively and quantitatively measure the quality of a position estimator?
\subsection{Datasets}
\noindent We focused on two standard robotic platforms of UAV flights and an inhouse dataset obtained with a terrestrial robot. UAVs with their 6 degrees of freedom are particularly suited and provide a challenging task for ego-motion estimation. 
\begin{itemize}
    \item \textbf{EuRoC MAV} \cite{burri_euroc_2016} is a dataset containing 6 drone flights in a small room. The drone has an on-board camera and inertial measurement units (IMU) for ego-pose estimation. This setup also includes MoCap.
    \item \textbf{Blackbird} \cite{antonini_blackbird_2020} contains 168 aggressive drone flights. The platform is also equipped with IMU and MoCap. ``Virtual'' onboard camera frames are available, but correspond to CG renderings in virtual environments rendered using the estimated odometry information.
    \item \textbf{LoCoBot-INSAL} is our inhouse dataset consisting of 3 trajectories of a terrestrial LoCoBot~\cite{locobot} navigating in the lab space at INSA-Lyon and ego-motion estimated with 3 sensors: IMU, wheel encoders, and a Lidar.
\end{itemize}
% En fait il y Z floor pour tout le monde : on compare toujours les deux approches. Ca permet de dire que depth depuis la z-cam et depth sur le floor donne les mêmes résultat, ce qui est bien
\noindent
We used VICE to manually label each frame using the tool shown in Fig.  \ref{fig:interface}. We labelled the three flights V1 for \textit{EuRoC MAV},  they present three different types of control from fairly slow to aggressive. For the \textit{Blackbird} dataset we labelled one portion of the flight in the environment Patrick with a constant yaw and a medium range speed for control. For \textit{LoCoBot-INSAL} dataset we labeled one section of one trajectory. 
%As the control of the robot if fairly slow the robot makes several pauses during the trajectory.

For all three datasets we used the method described in section \ref{sec:estimatingdepth} to estimate depth values, which we labelled ``\textit{Floor}'' in the tables of this section. For \textit{LoCoBot-INSAL}, a z-cam was installed on the robot. For \textit{EuRoC MAV} we provide an alternative way to calculate depths, which leverages the availability of a full 3D scan of the scenes. Following \cite{gordon_depth_2019}, we project the point cloud into the camera frame, and filter the points out of the field of view. The distance between the remaining points and the optical center of the camera gives a sparse image of the depths extended afterwards by linear interpolation. We down-sampled the video to $10$ FPS and crop them to $30$ seconds. Then, we manually labelled four reference keypoints across each sequence.
% AND THAT WAS EXCRUCIATING

\subsection{How to evaluate an evaluation method}
\noindent
VICE is a method which allows to evaluate ego-pose estimation methods, and which therefore uses performance metrics to produce its results. Our claim is that VICE is useful for this purpose, and we evaluate this claim by comparing VICE's output to precision measurements produced using ego-pose ground truth. Precise GT is unfortunately not available, which is why we compare with measurements obtained with marker based MoCap. We claim and argue that VICE produces results even more precise than MoCap, as can be seen in some examples in figure \ref{fig:3x7_vice}. However, given the sharp difference in quality to the noisy pose-estimation algorithms we evaluate, based on IMUs and SLAM, the comparison with MoCap data can serve as an approximate performance evaluation.

To this end, we compare the UAV ego-pose transforms $T^{\mathcal{B}_{t+1}, \mathcal{B}_t} = \left(d_t, \mathbf{R}_t\right)$ as measured by MoCap to the on-board measurements $\hat{T}^{\mathcal{B}_{t+1}, \mathcal{B}_t}$ to be evaluated using absolute AOE \cite{brossard_denoising_2020} and relative ROE \cite{zhang_tutorial_2018} angle errors:

%we compare the UAV position $\mathbf{x}$ and angles $\mathbf{R}$ as measured by MoCap to the measurements to be evaluated $(\hat{\mathbf{x}}, \hat{\mathbf{x}})$, using absolute AOE \cite{brossard_denoising_2020} and relative ROE \cite{zhang_tutorial_2018} angle errors:
% \begin{equation}
%     \text{RMSE}_{3D} = \sqrt{\frac{1}{T}\sum_{t=1}^T \|\mathbf{x}_t - \hat{\mathbf{x}}_t\|^2_2}.
% \end{equation}
\begin{equation}
    \text{RMSE}_{3D} = \sqrt{\frac{1}{T}\sum_{t=1}^T \|d_t - \hat{d}_t\|^2_2}.
\end{equation}
$$
\mathrm{AOE}=\sqrt{\sum_{t=1}^{T} \frac{1}{T}\left\|\log \left(\mathbf{R}_{t}^{T} \hat{\mathbf{R}}_{t}\right)\right\|_{2}^{2}}
$$
$$
\mathrm{ROE}=\left\|\log \left(\delta \mathbf{R}_{t, t+1}^{T} \delta \hat{\mathbf{R}}_{t, t+1}\right)\right\|_{2}
$$
%with $t$ a sample of the orientation. 
The ROE is uniformly sampled over time and averaged.

The above metrics measure GT ego-pose estimation performance and give an indication of the true error, but can't be compared to the performance as measured by VICE ( RMSE$_{2D}$, 
equation (\ref{eq:RMSE2D})). To this end, we introduce a metric, which directly evaluates the difference between the ground-truth error and the error as measured by VICE, namely RMSE$_{2D}$ as given in equation  (\ref{eq:RMSE2D}) between the reference keypoint annotated with VICE and the keypoint calculated using MoCap estimated poses, equation (\ref{eq:tracking}), which we call ``$\Delta$(VICE,GT)'' in this section.

% \textit{RMSE:}
% $$
% \sphericalangle = cos^{-1}\left(\frac{\mathbf{p^\mathcal{C}_\text{label}}}{\left\|\mathbf{p^\mathcal{C}_\text{label}}\right\|_{2}}.\frac{\mathbf{p^\mathcal{C}_\text{ego-pose}}}{\left\|\mathbf{p^\mathcal{C}_\text{ego-pose}}\right\|_{2}}\right)
% $$

% While not being absolutely necessary \sj{is that explained somewhere in the method ?}, VICE takes advantages of depth images from the on-board camera. Unfortunately, the drone used in EuRoC MAV does not integrate a depth camera. 

\begin{table*}
\centering
\definecolor{cadre}{gray}{0.9}
\renewcommand{\arraystretch}{1.1}
\NiceMatrixOptions{cell-space-limits = 0pt}
\begin{NiceTabular}{c|c|ccc||c|ccc|ccc}
\CodeBefore 
   \columncolor{cadre}{1,2,6}
   \rowcolor{cadre}{1,2} 
\Body
\toprule
\multicolumn{2}{c|}{\multirow{2}{*}{}} &
\multicolumn{3}{c||}{\textbf{--- $\Delta$ (On-Board, GT) ---}} &
\multirow{2}{*}{Depth} & 
\multicolumn{3}{c|}{\textbf{--- $\Delta$ (VICE, GT) ---}}    &
\multicolumn{3}{c}{\textbf{--- $\Delta$ (VICE, On-Board) ---}} \\

& & \shortstack{RMSE \\ {[}m{]}} & AOE & ROE &
& 1 point & \shortstack{RMSE \\ 2 points} & 3 points &
    1 point & \shortstack{RMSE \\ 2 points}  & 3 points \\ \midrule

\Block{6-1}{EuRoC} &
    \Block{2-1}{V1 Easy} & 
        \Block{2-1}{0.040} & \Block{2-1}{0.05} & \Block{2-1}{0.042} &
        Floor  & 8.5$\pm$0.8 & 8.6$\pm$0.4 & 8.6$\pm$0.3 & 33.2$\pm$5.6 & 33.5$\pm$3.0 & 33.6$\pm$1.7 \\
        & & & & & 
        Z-Map  & 8.6$\pm$0.7 & 8.7$\pm$0.4 & 8.7$\pm$0.2 & 34.4$\pm$4.5 & 34.6$\pm$2.5 & 34.7$\pm$1.4 \\ \cline{2-12}
&   \Block{2-1}{V2 Medium} &
        \Block{2-1}{0.138} & \Block{2-1}{0.10} & \Block{2-1}{0.109} &
        Floor & 11.8$\pm$0.8 & 11.8$\pm$0.4 & 11.8$\pm$0.3 & 44.9$\pm$4.1 & 45.1$\pm$2.4 & 45.1$\pm$1.4  \\ 
        & & & & &
        Z-Map & 16.0$\pm$6.1 & 16.6$\pm$4.1 & 17.0$\pm$2.5 & 46.0$\pm$5.3 & 46.2$\pm$3.1 & 46.3$\pm$1.8   \\ \cline{2-12}
&   \Block{2-1}{V3 Difficult} &
        \Block{2-1}{0.139} & \Block{2-1}{0.14} & \Block{2-1}{0.094} &
        Floor & 15.5$\pm$1.8 & 15.6$\pm$1.0 & 15.6$\pm$0.6 & 78.5$\pm$6.7 & 78.7$\pm$3.8 & 78.7$\pm$2.2  \\
        & & & & & 
        Z-Map & 24.7$\pm$10.6 & 26.0$\pm$6.8 & 26.6$\pm$4.0 & 79.9$\pm$6.9 & 80.1$\pm$3.8 & 80.2$\pm$2.2   \\\midrule
        
\Block{4-1}{\shortstack{LoCoBot\\INSAL}} & 
    \Block{2-1}{SLAM} &
        \multicolumn{3}{c||}{\multirow{2}{*}{N.A}} &
        Floor & \multicolumn{3}{c|}{\multirow{2}{*}{N.A}} & 46.2$\pm$5.6 & 46.4$\pm$3.4 & 46.5$\pm$2.0  \\
        & & \multicolumn{3}{c||}{} &
        Z-Map & \multicolumn{3}{c|}{} & 47.4$\pm$5.1 & 47.5$\pm$3.1 & 47.6$\pm$1.8  \\ \cline{2-12}
&   \Block{2-1}{Odom} &
        \multicolumn{3}{c||}{\multirow{2}{*}{N.A}} &
        Floor & \multicolumn{3}{c|}{\multirow{2}{*}{N.A}} & 14.4$\pm$4.4 & 14.8$\pm$2.9 & 15.0$\pm$1.8  \\
        & & \multicolumn{3}{c||}{} &
        Z-Map & \multicolumn{3}{c|}{} & 13.0$\pm$2.7 & 13.2$\pm$1.7 & 13.2$\pm$1.0   \\ \midrule
        
\Block{2-1}{Blackbird} &
    \Block{2-1}{Patrick} & 
        \Block{2-1}{0.215} & \Block{2-1}{0.01} & \Block{2-1}{0.014} & 
        Floor & 6.5$\pm$1.6 & 6.6$\pm$0.9 & 6.7$\pm$0.5 & 68.9$\pm$13.1 & 69.7$\pm$7.7 & 70.0$\pm$4.4 \\
        & & & & &
        Z-Map & 6.5$\pm$1.6 & 6.7$\pm$0.9 & 6.7$\pm$0.5 & 77.5$\pm$11.3 & 78.0$\pm$6.5 & 78.2$\pm$3.8 \\ \bottomrule
\end{NiceTabular}
\vspace*{2mm}
\caption{\label{tab:results}Quantitative results. Left block: comparison of onboard estimates with GT (real precision); middle block: difference between VICE and GT (calibration, i.e. evaluation of VICE); right block: VICE output, i.e. estimate of the precision of on-board estimates. Averages and standard deviations have been obtained by combining different set of labeled reference points}
\vspace{-5mm}
\end{table*}

\begin{figure*}[t]
    \centering
    \includegraphics[width=0.95\linewidth]{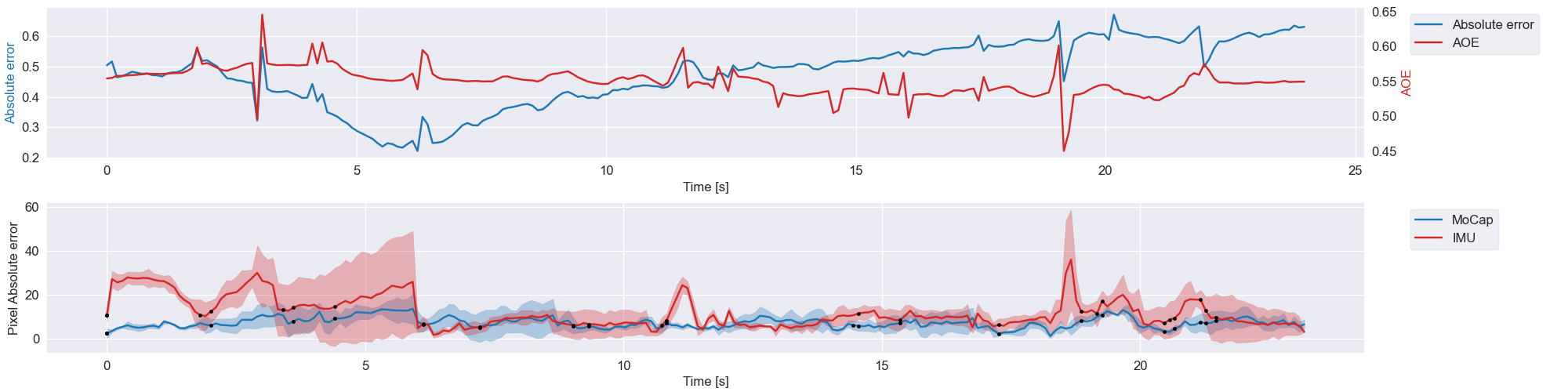}
    \caption{The graphs correspond to a section of V1 Easy from EuRoC MAV. The first graph shows the evolution of the AOE and the absolute error of position between the MoCap and the IMU data. The second graph describes the evolution of the absolute 2D error averaged over several tracked points. The black dots refers to the time where the algorithm resets \emph{i.e.} where the tracked point is reinitialized in the scene. 
    We observe a temporal correlation between the sensor error and the VICE error.}
    %\caption{The graphs correspond to a section of V1 Easy from EuRoC MAV. The first graph shows the evolution of the AOE and the absolute error of position between the MoCap and the IMU data. The second graph describes the evolution of the absolute 2D error averaged over several tracked points. The black dots refers to the time where the algorithm resets \emph{i.e.} where the tracked point is reinitialized in the scene. %Although relevant only for the second graph, the dotted lines are put on the first for reading purpose.}
    \label{fig:qualitative}
    \vspace{-5mm}
\end{figure*}

%The \emph{VICE} algorithm has been applied to the EuRoC MAV dataset \cite{burri_euroc_2016} using its motion capture for a first pose estimation and comparing it to the IMU.
%The depth map have been estimated form the point cloud.
%The flight V02 have not been synchronised ?
%Video from flight V01\_01\_easy \url{https://youtu.be/kbiO6G1pM-E}

%\subsection{Metrics}
% Pixel and 3d error

% Finally, because the 2D points describe an object in the scene with a particular depth, we also compute the RMSE between $\mathbf{p^\mathcal{C}_\text{label}}$ and $\mathbf{p^\mathcal{C}_\text{ego-pose}}$

\subsection{Results}
\noindent
% The main functionality of our tool is to rapidly and intuitively evaluate the overall quality of the data gathering pipeline. As a demonstration, 
We compute the RMSE on the flights from EuRoC MAV and Blackbird datasets using two different sensors for pose estimation : (a) a noisy visual-inertial odometry (OpenVINS \cite{geneva_openvins_2020}) for EuRoC MAV and  Euler integration scheme for Blackbird and (b) a more precise external pose estimation from a MoCap device.
Moreover for the LoCoBot INSAL dataset, we compute the metrics for two ego-pose estimation methods: one is based on an odometry - \emph{Odom} - and the other is obtained using \emph{SLAM} \cite{SLAMROS2021}.

\myParagraph{Qualitative evaluation} 
Figure \ref{fig:3x7_vice} shows example video frames with superimposed 2D keypoints as estimated by (i) VICE, (ii) the GT measurement (MoCap), and (iii) the on-board ego-pose estimation system, based on IMUs, encoders, or SLAM. As we can see, VICE keypoints as annotated by humans closely track semantic points in the scene (e.g. a point on a chair), whereas even keypoints placed by MoCap show error. This indicates, that MoCap is not the most precise way to evaluate pose-estimation, even though it is widely used for the creation of GT measurements. We can also see that the precision of the on-board pose estimations (green dots) is significantly lower.
We also provide videos\footnote{\url{https://youtube.com/playlist?list=PLRsYEUUGzW54jqsfRdkNAYjZUnoEM4uhM}}, where we  observe significant differences from \textit{easy} to \textit{difficult} flights. 

\myParagraph{Quantitative evaluation} %VICE directly assess the precision of the pose estimationQualitative evaluation the difference between the human-labelled keypoint and its estimation from the pose measurements. 
Table \ref{tab:results} reports results using all proposed metrics, and provides evidence that the pose is better estimated using MoCap, compared to onboard systems, which does not come as a surprise. We can see that precision decreases with the aggressiveness of the flight in \textit{EuRoC MAV}. Noticeably, a unique labeled point is enough for VICE to measure the accuracy of an ego-pose estimate. Increasing the number of point yet reduces the dependency on (arbitrary) choice of initial position. 
%Moreover as the ego-pose becomes more and more precise the metrics of the algorithm tends towards 0 as shown with the MoCap of blackbird. Finally theses results are qualitatively observable in the videos.
%We confirm these results by comparing numerically the difference between them with several datasets. 

\myParagraph{Calibration: evaluating evaluation}
The middle block in table \ref{tab:results} provides an evaluation of the VICE method as a difference between the pose error it estimates and the pose error estimated through MoCap. This error is small, in particular compared to the error of the onboard estimate itself (right block). We can also see that the error is lower for the easier flight than for the difficult ones, and this order correlates with the GT error (left block).

\myParagraph{Depth} Table \ref{tab:results} compares  results between the floor method of depth estimation (c.f. section \ref{sec:estimatingdepth}) using z-Cam. We note  similar measurements, with the exception of the aggressive flight, which we conjecture is due to alignment problems in these difficult high-motion conditions. This confirms the applicability of the floor method in situations where no depth is available.

\section{Conclusion}
\noindent
We introduced a new method and tool for evaluating the precision of UAV/robot odometry algorithms. Our approach aims to be simple to implement and does away with expensive equipment (MoCap, Lidar, etc.), obtaining GT measurements through computer vision in a semi-automatic way by casting the task as 2D point tracking problem. We have proposed different experiments highlighting the potential of VICE in the use of robotic platforms dedicated to computer vision.

\bibliographystyle{IEEEtran}
\bibliography{references, refs_chris}

\begin{thebibliography}{10}
\providecommand{\url}[1]{#1}
\csname url@samestyle\endcsname
\providecommand{\newblock}{\relax}
\providecommand{\bibinfo}[2]{#2}
\providecommand{\BIBentrySTDinterwordspacing}{\spaceskip=0pt\relax}
\providecommand{\BIBentryALTinterwordstretchfactor}{4}
\providecommand{\BIBentryALTinterwordspacing}{\spaceskip=\fontdimen2\font plus
\BIBentryALTinterwordstretchfactor\fontdimen3\font minus
  \fontdimen4\font\relax}
\providecommand{\BIBforeignlanguage}[2]{{%
\expandafter\ifx\csname l@#1\endcsname\relax
\typeout{** WARNING: IEEEtran.bst: No hyphenation pattern has been}%
\typeout{** loaded for the language `#1'. Using the pattern for}%
\typeout{** the default language instead.}%
\else
\language=\csname l@#1\endcsname
\fi
#2}}
\providecommand{\BIBdecl}{\relax}
\BIBdecl

\bibitem{AssemExpStudyArxiv2021}
A.~Sadek, G.~Bono, B.~Chidlovskii, and C.~Wolf, ``{An in-depth experimental
  study of sensor usage and visual reasoning of robots navigating in real
  environments.}'' in \emph{pre-print arXiv:2111.14666}, 2021.

\bibitem{revaud_r2d2_2019}
J.~Revaud, C.~De~Souza, M.~Humenberger, and P.~Weinzaepfel, ``{R2D2}:
  {Reliable} and {Repeatable} {Detector} and {Descriptor},'' \emph{Neural
  Information Processing Systems}, 2019.

\bibitem{geneva_openvins_2020}
P.~Geneva, K.~Eckenhoff, W.~Lee, Y.~Yang, and G.~Huang, ``{OpenVINS}: {A}
  {Research} {Platform} for {Visual}-{Inertial} {Estimation},'' \emph{IEEE
  International Conference on Robotics and Automation}, May 2020.

\bibitem{liu_tlio_2020}
W.~Liu, D.~Caruso, E.~Ilg, J.~Dong, A.~Mourikis, K.~Daniilidis, V.~Kumar,
  J.~Engel, A.~Valada, and T.~Asfour, ``{TLIO}: {Tight} {Learned} {Inertial}
  {Odometry},'' \emph{IEEE Robotics and Automation Letters}, Jul. 2020.

\bibitem{nisar_vimo_2019}
B.~Nisar, P.~Foehn, D.~Falanga, and D.~Scaramuzza, ``{VIMO}: {Simultaneous}
  {Visual} {Inertial} {Model}-{Based} {Odometry} and {Force} {Estimation},''
  \emph{IEEE Robotics and Automation Letters}, May 2019.

\bibitem{chen_ionet_2018}
C.~Chen, X.~Lu, A.~Markham, and N.~Trigoni, ``{IONet}: {Learning} to {Cure} the
  {Curse} of {Drift} in {Inertial} {Odometry},'' \emph{AAAI Conference on
  Artificial Intelligence}, Jan. 2018.

\bibitem{cadena_simultaneous_2016}
C.~Cadena, L.~Carlone, H.~Carrillo, Y.~Latif, D.~Scaramuzza, J.~Neira, I.~Reid,
  and J.~Leonard, ``Simultaneous {Localization} {And} {Mapping}: {Present},
  {Future}, and the {Robust}-{Perception} {Age},'' \emph{IEEE Transactions on
  Robotics}, Jun. 2016.

\bibitem{engel_lsd-slam_2014}
J.~Engel, T.~Schoeps, and D.~Cremers, ``{LSD}-{SLAM}: {Large}-scale direct
  monocular {SLAM},'' \emph{European Conference on Computer Vision}, Sep. 2014.

\bibitem{mur-artal_orb-slam_2015}
R.~Mur-Artal, J.~Montiel, and J.~Tardos, ``{ORB}-{SLAM}: a versatile and
  accurate monocular {SLAM} system,'' \emph{IEEE Transactions on Robotics},
  Oct. 2015.

\bibitem{SLAMROS2021}
J.~I. Macenski, S., ``{SLAM Toolbox: SLAM for the dynamic world},''
  \emph{{Journal of Open Source Software}}, vol.~6, no.~61, 2011.

\bibitem{newcombe_kinectfusion_2011}
R.~Newcombe, A.~Davison, S.~Izadi, P.~Kohli, O.~Hilliges, J.~Shotton,
  D.~Molyneaux, S.~Hodges, D.~Kim, and A.~Fitzgibbon, ``{KinectFusion}:
  {Real}-time dense surface mapping and tracking,'' \emph{IEEE International
  Symposium on Mixed and Augmented Reality}, Oct. 2011.

\bibitem{TopoSLAM2001}
H.~Choset and K.~Nagatani, ``{Topological simultaneous localization and mapping
  (SLAM): toward exact localization without explicit localization},''
  \emph{{IEEE Transactions on Robotics and Automation}}, vol.~17, no.~2, 2001.

\bibitem{forsyth_computer_2011}
D.~Forsyth and J.~Ponce, \emph{Computer {Vision}: {A} {Modern} {Approach}.
  ({Second} edition)}.\hskip 1em plus 0.5em minus 0.4em\relax Prentice Hall,
  Nov. 2011.

\bibitem{SuperGLUECVPR202}
P.~Sarlin, D.~DeTone, T.~Malisiewicz, and A.~Rabinovich, ``Superglue: Learning
  feature matching with graph neural networks,'' in \emph{Computer Vision and
  Pattern Recognition (CVPR)}, 2020.

\bibitem{fridman_automated_2015}
L.~Fridman, D.~Brown, W.~Angell, I.~Abdić, B.~Reimer, and H.~Y. Noh,
  ``Automated {Synchronization} of {Driving} {Data} {Using} {Vibration} and
  {Steering} {Events},'' \emph{Pattern Recognition Letters}, Oct. 2015.

\bibitem{ploetz_automatic_2012}
T.~Ploetz, C.~Chen, N.~Hammerla, and G.~Abowd, ``Automatic {Synchronization} of
  {Wearable} {Sensors} and {Video}-{Cameras} for {Ground} {Truth} {Annotation}
  –- {A} {Practical} {Approach},'' \emph{International Symposium on Wearable
  Computers}, Jun. 2012.

\bibitem{zhang_syncwise_2020}
Y.~Zhang, S.~Zhang, M.~Liu, E.~Daly, S.~Battalio, S.~Kumar, B.~Spring, J.~Rehg,
  and N.~Alshurafa, ``{SyncWISE}: {Window} {Induced} {Shift} {Estimation} for
  {Synchronization} of {Video} and {Accelerometry} from {Wearable} {Sensors},''
  \emph{Interactive, Mobile, Wearable and Ubiquitous Technologies}, Sep. 2020.

\bibitem{rehder_extending_2016}
J.~Rehder, J.~Nikolic, T.~Schneider, T.~Hinzmann, and R.~Siegwart, ``Extending
  kalibr: {Calibrating} the extrinsics of multiple {IMUs} and of individual
  axes,'' \emph{IEEE International Conference on Robotics and Automation}, May
  2016.

\bibitem{brossard_denoising_2020}
M.~Brossard, S.~Bonnabel, and A.~Barrau, ``Denoising {IMU} {Gyroscopes} with
  {Deep} {Learning} for {Open}-{Loop} {Attitude} {Estimation},'' \emph{IEEE
  Robotics and Automation Letters}, Jun. 2020.

\bibitem{zhang_tutorial_2018}
Z.~Zhang and D.~Scaramuzza, ``A {Tutorial} on {Quantitative} {Trajectory}
  {Evaluation} for {Visual}(-{Inertial}) {Odometry},'' \emph{IEEE International
  Conference on Intelligent Robots and Systems}, pp. 7244--7251, Oct. 2018.

\bibitem{Brown1966DecenteringDO}
D.~Brown, ``Decentering distortion of lenses,'' in \emph{Photogrammetric
  Engineerings}, 1966.

\bibitem{burri_euroc_2016}
M.~Burri, J.~Nikolic, P.~Gohl, T.~Schneider, J.~Rehder, S.~Omari, M.~Achtelik,
  and R.~Siegwart, ``The {EuRoC} micro aerial vehicle datasets,'' \emph{The
  International Journal of Robotics Research}, Jan. 2016.

\bibitem{antonini_blackbird_2020}
A.~Antonini, W.~Guerra, V.~Murali, T.~Sayre-McCord, and S.~Karaman, ``The
  {Blackbird} {Dataset}: {A} {Large}-{Scale} {Dataset} for {UAV} {Perception}
  in {Aggressive} {Flight},'' \emph{International Symposium on Experimental
  Robotics}, Jan. 2020.

\bibitem{locobot}
``{LoCoBot: An Open Source Low Cost Robot},'' \url{http://www.locobot.org}.

\bibitem{gordon_depth_2019}
A.~Gordon, H.~Li, R.~Jonschkowski, and A.~Angelova, ``Depth {From} {Videos} in
  the {Wild}: {Unsupervised} {Monocular} {Depth} {Learning} {From} {Unknown}
  {Cameras},'' \emph{International Conference on Computer Vision}, Oct. 2019.

\end{thebibliography}

\end{document}